\documentclass[letterpaper, 10 pt, conference]{ieeeconf}

\IEEEoverridecommandlockouts
\usepackage{cite}
\usepackage{amsmath,amssymb,amsfonts,mathtools}
\usepackage{graphicx}
\usepackage{textcomp}
\usepackage{xcolor}
\usepackage{bm}
\usepackage{algorithm}
\usepackage{adjustbox}
\usepackage{algpseudocode}
\def\BibTeX{{\rm B\kern-.05em{\sc i\kern-.025em b}\kern-.08em
    T\kern-.1667em\lower.7ex\hbox{E}\kern-.125emX}}

\usepackage{comment}
\usepackage{subcaption}
\usepackage{caption}
\usepackage{MnSymbol}
\usepackage{dsfont}
\usepackage{pdfpages}
\usepackage{booktabs}
\usepackage{adjustbox}
\usepackage{multirow}

\usepackage{amsthm}

\newtheorem{theorem}{Theorem}
\newtheorem{lemma}{Lemma}
\newtheorem{remark}{Remark}

\DeclareMathOperator{\PPC}{PPC}

\makeatletter
\newcommand\thefontsize{\f@size pt}
\makeatother
 
\begin{document}

\title{\LARGE \bf Robust Point Cloud Reinforcement Learning via PCA-Based Canonicalization}

\author{%
  Michael Bezick$^{1}$, Vittorio Giammarino$^{1}$, Ahmed H. Qureshi$^{1}$%
  \thanks{$^{1}$Department of Computer Science, Purdue University, West Lafayette, IN 47907, USA.
  {\tt\small \{mbezick, vgiammar, ahqureshi\}@purdue.edu}}%
}

\maketitle

\begin{abstract}
    Reinforcement Learning (RL) from raw visual input has achieved impressive successes in recent years, yet it remains fragile to out-of-distribution variations such as changes in lighting, color, and viewpoint. Point Cloud Reinforcement Learning (PC-RL) offers a promising alternative by mitigating appearance-based brittleness, but its sensitivity to camera pose mismatches continues to undermine reliability in realistic settings. To address this challenge, we propose PCA Point Cloud (PPC), a canonicalization framework specifically tailored for downstream robotic control. PPC maps point clouds under arbitrary rigid-body transformations to a unique canonical pose, aligning observations to a consistent frame, thereby substantially decreasing viewpoint-induced inconsistencies. In our experiments, we show that PPC improves robustness to unseen camera poses across challenging robotic tasks, providing a principled alternative to domain randomization.
\end{abstract}


\section{Introduction}

\begin{figure*}[t]
    \centering
    \smallskip
    \smallskip
    \adjustbox{width =1\textwidth} {
    \includegraphics{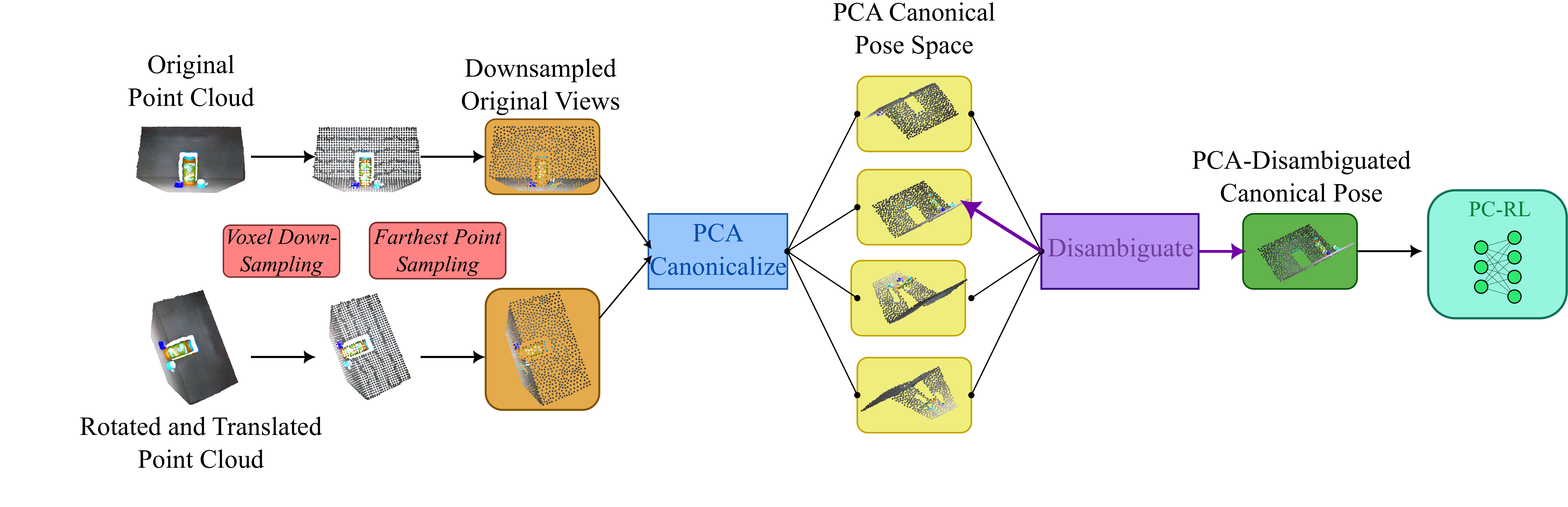}}
    \caption{Illustration of the proposed \textit{PCA Point Cloud (PPC)} pipeline. Given an input point cloud, PPC first applies \textit{voxel downsampling} to regularize point distribution, followed by \textit{Farthest Point Sampling (FPS)} to select a fixed number of representative points, ensuring consistent shape coverage and enabling uniform batch processing. A PCA-based canonicalization is then performed, producing four candidate eigenvector bases, from which a geometry-driven disambiguation step deterministically selects one. The point cloud is subsequently re-expressed in this basis, yielding a unique disambiguated canonical pose. As a result, regardless of the initial rotation or translation of the input, the final canonical representation remains consistent.}
    \vspace{-0.2cm}
    \label{fig:pca_diambiguation}
\end{figure*}

Reinforcement Learning (RL) from pixels, often referred to as Visual RL, aims to learn policies end-to-end by mapping high-dimensional image observations directly to actions. This paradigm has recently attracted considerable attention in robotics as it allows agents to acquire complex behaviors from sensory streams such as camera images or video sequences, removing the need for hand-crafted perception pipelines~\cite{mnih2015human, pinto2018asymmetric, chi2023diffusion, geles2024demonstrating}. Yet the same reliance on raw visual input also exposes a critical weakness: pixel-based policies are notoriously brittle to distribution shifts, where even small changes in lighting, background, or camera pose can lead to severe performance degradation~\cite{yuan2023rl, giammarino2024visually, ramazzina2025beyond}. As a result, despite strong in-distribution performance, their lack of robustness to visual variations remains a major obstacle to real-world deployment.

To overcome these limitations, recent work has turned to Point Cloud Reinforcement Learning (PC-RL)~\cite{gyenes2024pointpatchrl}, which defines policies directly on 3D point cloud observations. By leveraging geometric representations, PC-RL reduces sensitivity to nuisance factors such as lighting, background, and field of view~\cite{peri_point_2024}, and has proven particularly effective for tasks that require fine-grained reasoning about robot-object relationships~\cite{ling2023efficacy}. Despite these advantages, point clouds are typically expressed in a camera's local coordinate frame, making them inherently sensitive to sensor pose. Even modest translations or rotations can drastically change the scene representation, leading to instability and degraded performance at deployment. Consequently, while PC-RL alleviates appearance-related brittleness, it continues to suffer from viewpoint sensitivity.

A common strategy to address this problem is given by domain randomization, where training environments are deliberately perturbed to encourage robustness. In the context of PC-RL, this typically involves applying random rotations and translations to point clouds so that policies are trained across a wide range of poses. By aiming to generalize across a vast distribution of perturbations, domain randomization often leads to substantial sample inefficiency. In addition, its theoretical guarantees are weak and rely on assumptions that are rarely satisfied in practice~\cite{tobin2017domain, mehta2020active}.

In this work, we present a principled alternative to domain randomization that tackles PC-RL’s viewpoint sensitivity both effectively and efficiently. Drawing on advances in rotation-invariant point cloud analysis~\cite{li2021closer, luo2024general}, we introduce PCA Point Cloud (PPC), a canonicalization technique based on Principal Component Analysis (PCA) that standardizes a point cloud's reference frame, producing one consistent representation. Specifically, PPC aligns each point cloud with its principal axes and resolves PCA’s inherent sign ambiguity through a novel geometry-driven disambiguation step, ensuring that every input is mapped to a unique canonical pose. This guarantees invariance to both translation and rotation, effectively neutralizing PC-RL’s sensitivity to coordinate frames. We show that this simple yet rigorous procedure substantially improves the robustness of PointPatch RL (PPRL)~\cite{gyenes2024pointpatchrl}, enabling it to maintain strong performance under viewpoint shifts where the vanilla algorithm frequently collapses. We summarize our main contributions as follows:
\begin{itemize}
\item We propose \textit{PCA Point Cloud (PPC)}, a novel canonicalization method that maps each point cloud to a unique pose, guaranteeing invariance to rigid transformations.
\item We integrate PPC into PPRL~\cite{gyenes2024pointpatchrl} and demonstrate, through extensive experiments, substantial gains in robustness to unseen camera viewpoints across diverse robotic tasks.
\item We conduct additional tests on real-world point clouds using PPC to demonstrate how point cloud pose invariance effectively approximates camera pose invariance.
\end{itemize}

\section{Related Work}

\paragraph{Point Cloud Representation Learning} 

A central challenge in enabling end-to-end control from high-dimensional perceptual inputs is the design of robust representation learning pipelines that compress raw data into compact but informative embeddings. Such representations have underpinned the success of both model-free~\cite{mnih2015human, giammarinoadversarial} and model-based~\cite{hafner2019learning} Visual RL. In the context of point clouds, early work focused on learning directly from unordered sets. PointNet~\cite{qi2017pointnet} introduced permutation invariance via shared per-point MLPs and symmetric pooling, but struggled to capture local geometric structure. PointNet++~\cite{qi2017pointnet++} improved upon this via hierarchical grouping and Farthest Point Sampling (FPS) to model local neighborhoods, though it remained sensitive to sampling density, scale, and rotations. Building on these foundations, PointTransformer~\cite{Zhao_2021_ICCV} introduced vector attention mechanisms to capture interactions within local neighborhoods, while Point-MAE~\cite{doi:10.1142/S2811032324400010} adapted masked autoencoding by reconstructing randomly masked patches with a Chamfer distance loss. Point-BERT~\cite{Yu_2022_CVPR} replaced random masking with block masking and discrete token prediction, and Point-GPT~\cite{chen2023pointgptautoregressivelygenerativepretraining} further extended this direction by adopting autoregressive inference strategies. Despite these advances, none of these methods provide robustness to rigid transformations such as translations or rotations, which remains a central challenge for their application to real-world control.

\paragraph{Point Cloud Reinforcement Learning} 

Existing work in PC-RL can be broadly categorized into: $(i)$ approaches leveraging pretrained encoders, $(ii)$ comparative studies against Visual RL, and $(iii)$ fully end-to-end PC-RL methods. Early efforts explored pretrained PointNet features for manipulation without full end-to-end optimization~\cite{huang2021generalization}, while others adopted 3D CNNs with distillation~\cite{chen2022system} or integrated pretrained point cloud encoders into imitation learning pipelines, where finetuning occurred during the imitation step~\cite{wu2023learning}. Beyond encoder design, Liu et al.~\cite{liu2023frame} showed that the choice of coordinate frame can significantly affect policy performance in PC-RL. Comparative evaluations further demonstrated that PC-RL policies can outperform Visual RL in mixed 2D/3D environments~\cite{ling2023efficacy}, underscoring the promise of 3D representations for control. More recently, specialized methods such as DexPoint~\cite{qin2023dexpoint} have advanced PC-RL for manipulation, while PPRL~\cite{gyenes2024pointpatchrl} introduced a patch-based encoder with auxiliary reconstruction objectives, establishing the state-of-the-art in PC-RL. All these existing methods remain sensitive to viewpoint mismatch, a critical limitation that motivates this work.

\paragraph{Point Cloud Rotation and Translation Invariance}

The problem of achieving invariance to rigid transformations in point clouds has been studied in recent years, though mostly outside the RL domain. Early approaches relied on learned orientation predictors~\cite{fang2020rotpredictor} or handcrafted geometric descriptors such as pairwise distances and angles~\cite{zhang2019rotation}. Orientation predictors are imperfect under distribution shift, while handcrafted descriptors often discard shape information critical for downstream control. Other methods enforce invariance by arranging feature maps into sorted or symmetric orders~\cite{xu2021sgmnet, kim2020rotation}, which removes orientation dependence but oversimplifies geometry by erasing relative spatial relationships between points. An alternative to these approaches is represented by PCA-based canonicalization, which proposes to align point clouds with the eigenvectors of their covariance matrix in order to define a canonical coordinate system~\cite{li2021closer, luo2024general}. While PCA-based methods can guarantee invariance to rigid transformations, they suffer from eigenvector sign ambiguities that yield multiple valid canonical poses. Recent work addresses this issue by restricting the canonical space to four orientations and blending features across them~\cite{luo2024general}, but this approach is unnecessarily costly when a single deterministic orientation is achievable in the absence of symmetries with respect to the principal axes. In contrast, we propose a lightweight procedure that enforces translation and rotation invariance by mapping any transformed point cloud to the same canonical pose while preserving its full geometric content and structure. Our method resolves PCA’s sign ambiguity through a geometry-driven disambiguation step, producing a unique and consistent pose that can be seamlessly integrated into any PC-RL algorithm.

\section{Preliminaries}
\label{sec:preliminaries}

\paragraph{Partially Observable Markov Decision Process} 

We model the decision process as an infinite-horizon discounted Partially Observable Markov Decision Process (POMDP) described by the tuple $(\mathcal{S}, \mathcal{A}, \mathcal{X}, \mathcal{T}, \mathcal{U}, \mathcal{R}, \rho_0, \gamma)$, where $\mathcal{S}$ is the set of states, $\mathcal{A}$ is the set of actions, and $\mathcal{X}$ is the set of observations. $\mathcal{T}:\mathcal{S}\times \mathcal{A} \rightarrow \mathcal{P}(\mathcal{S})$ is the transition probability function where $\mathcal{P}(\mathcal{S})$ denotes the space of probability distributions over $\mathcal{S}$, $\mathcal{U}:\mathcal{S} \rightarrow \mathcal{P}(\mathcal{X})$ is the observation probability function, and $\mathcal{R}:\mathcal{S}\times \mathcal{A} \rightarrow \mathbb{R}$ is the reward function which maps state-action pairs to scalar rewards. Finally, $\rho_0 \in \mathcal{P}(\mathcal{S})$ is the initial state distribution and $\gamma \in [0,1)$ the discount factor. In this framework, the true environment state $s \in \mathcal{S}$ is considered unobserved or partially observed by the agent. Given an action $a\in\mathcal{A}$, the next state is sampled such that $s'\sim\mathcal{T}(\cdot|s,a)$, a point cloud observation is generated as $x'\sim\mathcal{U}(\cdot|s')$, and a reward $\mathcal{R}(s,a)$ is computed. 

\paragraph{Reinforcement Learning} 

Given a POMDP and a stationary policy $\pi:\mathcal{S} \to \mathcal{P}(\mathcal{A})$, the RL objective is to maximize the expected discounted return $J(\pi)=\mathbb{E}_{\tau}\left[\sum_{t=0}^{\infty}\gamma^t \mathcal{R}(s_t,a_t)\right]$, where $\tau=(s_0,a_0,s_1,a_1,\dots)$ are trajectories obtained through interactions with the environment. When the underlying state $s_t$ is not fully observable, the policy is defined over a history of past observations, i.e., $\pi:\mathcal{X}^d \to \mathcal{P}(\mathcal{A})$. Yet the reward function $\mathcal{R}$ remains specified over the true state space $\mathcal{S}$ and the RL objective is to find $\pi$ such that $J(\pi)$ is maximized without fully observing $s_t$. To this end, RL algorithms typically rely on two auxiliary functions: the state value function $V^{\pi}(s)=\mathbb{E}_{\tau}\left[\sum_{t=0}^{\infty}\gamma^t \mathcal{R}(s_t,a_t)\big|S_0=s\right]$ and the state-action value function $Q^{\pi}(s,a)=\mathbb{E}_{\tau}\left[\sum_{t=0}^{\infty}\gamma^t \mathcal{R}(s_t,a_t)\big|S_0=s, A_0=a\right]$. These functions capture the expected cumulative reward when rolling out policy $\pi$ from a given initial state or state-action pair.

\paragraph{Point Patch Reinforcement Learning} 

PPRL~\cite{gyenes2024pointpatchrl} consists of three main components: $(i)$ a PointGPT-inspired transformer encoder for point cloud processing~\cite{chen2023pointgptautoregressivelygenerativepretraining}, $(ii)$ an actor-critic algorithm operating on the learned embedding~\cite{konda1999actor, haarnoja2018soft}, and $(iii)$ a masked reconstruction objective that strengthens representation learning. 
At each timestep, PPRL takes as input a point cloud $x_t \in \mathbb{R}^{n_t \times 3}$, where $n_t$ are the number of 3D points in the point cloud observed at time $t$. Note that we write $n_t$ with $t$ as subscript as this number can vary across time steps. Following Qi et al.~\cite{qi2017pointnet++}, $m$ centroids $c_t \in \mathbb{R}^{m \times 3}$ are sampled via FPS, and each centroid is used to define a local neighborhood through $k$-Nearest Neighbors, forming overlapping patches $P_t \in \mathbb{R}^{m \times k \times 3}$. Each patch is normalized relative to its centroid and embedded with a lightweight PointNet-style MLP~\cite{qi2017pointnet}, producing tokens $T_t \in \mathbb{R}^{m \times D}$. After adding sinusoidal positional encodings, the tokens are processed by a transformer encoder, and a sequence pooling layer~\cite{hassani2021escaping} yields a compact feature vector $z_t \in \mathcal{Z} \subset \mathbb{R}^l$. The actor-critic agent is then defined to operate entirely in this latent space, i.e., $\pi: \mathcal{Z} \to \mathcal{P}(\mathcal{A})$ and $Q^{\pi}: \mathcal{Z} \times \mathcal{A} \to \mathbb{R}$. Notably, actor and critic share the same point cloud encoder, which is updated end-to-end only via gradients from the critic loss as common practice in Visual RL~\cite{giammarinoadversarial}. To further enrich the feature vector $z_t$, PPRL incorporates a masked reconstruction objective inspired by PointGPT~\cite{chen2023pointgptautoregressivelygenerativepretraining}. After tokenization, patches are ordered along a Z-order curve, and the decoder receives encodings of the relative directions between successive patches. This design prevents trivial recovery of global structure and compels the model to exploit geometric context. Masking is applied both randomly and causally, and the transformer decoder is trained to reconstruct the missing patches by minimizing the Chamfer distance to the ground-truth patches. This reconstruction loss provides an additional geometric signal that improves the quality of $z_t$.

\section{Method}
\label{sec:methos}
In the following section, we present PPC, our PCA-based canonicalization pipeline designed to enhance PC-RL under viewpoint mismatch. A schematic overview of our approach is provided in Fig.~\ref{fig:pca_diambiguation}. The section is organized into three main parts. First, Section~\ref{subsec:downsampling} describes the point cloud downsampling steps that prepare the data for PCA-based canonicalization. Next, Section~\ref{subsec:PPC} formally introduces how PCA is applied within PPC to guarantee invariance of point clouds under any rigid transformation, along with theoretical results that support this claim. Finally, Section~\ref{subsec:PPC+PPRL} summarizes the complete PPC pipeline and its integration into PC-RL.

\subsection{Point Cloud Downsampling}
\label{subsec:downsampling}

In PC-RL the raw observation at each time step is represented by a point cloud $x_t \in \mathbb{R}^{n_t \times 3}$, where $n_t$ denotes the number of 3D points observed at time $t$. Since $n_t$ varies with changes in camera pose and field of view, preprocessing is required to obtain a consistent representation. To this end, we propose a two-stage downsampling pipeline comprised of \emph{voxel downsampling} followed by \emph{farthest point sampling (FPS)}. The voxel stage enforces a minimum spatial separation between points, efficiently removing redundancies and equalizing density without heavy averaging~\cite{rusu20113d}, while also reducing the computational load for the subsequent stage. FPS is then applied to the voxel-thinned cloud to select exactly $n$ points, ensuring balanced spatial coverage of the visible geometry compared to random subsampling~\cite{qi2017pointnet++}. For simplicity, we continue to denote the resulting fixed-size point cloud as $x_t \in \mathbb{R}^{n \times 3}$ where $n$ drops the subscript.  

This design combines the speed and density regularization of voxel downsampling with the coverage-oriented selection of FPS, yielding $(i)$ a fixed-size input that simplifies batching, $(ii)$ smoother representations under small camera pose changes (less grid-boundary snapping than pure voxel grids), and $(iii)$ more stable centroids and PCA axes for canonicalization, since they are computed from a coverage-balanced subset. This fixed-size cloud serves as the input to our PCA-based canonicalization step and provides improved robustness to viewpoint changes.

\subsection{PCA Point Cloud}
\label{subsec:PPC}

We now introduce the core of our PPC method, i.e., our PCA-based canonicalization step, which enables invariance to any rigid transformation of the point cloud. This is crucial in PC-RL with mismatch, where training occurs with a given camera pose, so that $x_t \sim \mathcal{U}(\cdot|s_t)$, while evaluation occurs with unseen poses, i.e., $\widetilde x_t \sim \widetilde{\mathcal{U}}(\cdot|s_t)$ for the same state $s_t$ where $x_t \neq \tilde x_t$. This problem can be addressed by means of domain randomization, which broadens the training distribution to cover possible evaluation poses. However, randomization seeks generalization in the embedding space and can be particularly sample-inefficient and ineffective when mismatches are large.  

Our approach takes a complementary route by formulating a point cloud canonicalization map $\mathcal{C}$ with four properties: 
\begin{itemize}
    \item \textbf{Pose invariance}: for any rotation matrix $\bm R\!\in\!SO(3)$ and translation vector $\bm \iota \!\in\!\mathbb{R}^{3 \times 1}$, $\mathcal{C}(x \bm R^{\top} + \bm 1_n \bm \iota^{\top}) = \mathcal{C}(x)$ where $\bm 1_n \in \mathbb{R}^{n \times 1}$ is a column vector of ones;  
    
    \item \textbf{Shape preservation}: rigid distances are preserved so control signals remain faithful;  
    
    \item \textbf{Determinism}: each input maps to a unique output;  
    
    \item \textbf{Constant dimensionality}: the representation remains $n \times 3$ without inflated feature size. 
\end{itemize}  

Formally, for any point cloud $x \!\in\! \mathbb{R}^{n \times 3}$ and rigid transform $(\bm R,\bm \iota)$, given $x' = x \bm R^{\top} + \bm 1_n \bm \iota^\top$ we guarantee that $\mathrm{PPC}(x') \equiv \mathrm{PPC}(x)$ under the simple-spectrum and non-degenerate $\phi$ conditions (see Theorem \ref{thm:rigid_invariance}). The resulting output becomes a \emph{unique}, translation- and rotation-invariant canonical pose that preserves all geometric detail without increasing dimensionality. We next overview the canonicalization method.

\paragraph{Overview}
\label{sec:overview}
Given a fixed-size cloud $x \in \mathbb{R}^{n \times 3}$, rigid transformation invariance is achieved through three main steps: $(i)$ centering the point cloud for translation invariance, $(ii)$ aligning to the eigenvectors of the covariance matrix to remove rotation, and $(iii)$ resolving PCA sign ambiguity with a geometry-driven disambiguation function $\phi$.

Specifically, we start by defining the centroid of a point cloud as $\mu(x) = \frac{1}{n}\sum_i x(i)$,
where we abuse notation and write $x(i) \in \mathbb{R}^{1\times 3}$ for the $i$-th point of $x$. Translation is removed by subtracting $\mu(x)$ from each point, i.e., $\bar{x}(i) = x(i) - \mu(x)$, yielding the centered cloud $\bar{x} \in \mathbb{R}^{n \times 3}$.

The PCA-based canonicalization step follows by defining the eigendecomposition
\begin{equation}
    \Sigma(\bar{x}) = \frac{1}{n-1}\bar{x}^\top \bar{x} = \bm E(\bar{x}) \bm \Lambda(\bar{x}) \bm E^\top(\bar{x}),
    \label{eq:PCA}
\end{equation}
where $\bm \Lambda(\bar{x}) = \text{diag}(\lambda_1,\lambda_2,\lambda_3)$ contains the eigenvalues and $\bm E(\bar{x}) = [\bm e_1\ \bm e_2\ \bm e_3] \in \mathbb{R}^{3 \times 3}$ the corresponding eigenvectors. For a rotated cloud $\bar{x}' = \bar{x}\bm R^\top$, the covariance in \eqref{eq:PCA} becomes $\Sigma(\bar{x}') = \bm R \Sigma(\bar{x}) \bm R^\top$, whose eigenbasis is $\bm E(\bar{x}') = \bm R \bm E(\bar{x})$ (see Lemma~\ref{lemma:1}). Thus,
\begin{equation*}
    \bar{x}' \bm E(\bar{x}') = (\bar{x} \bm R^\top)(\bm R \, \bm E(\bar{x})) = \bar{x} \bm E(\bar{x}),
\end{equation*}
showing rotation invariance with respect to the canonical pose induced by the eigenbasis $\bm E(\cdot)$.

However, as emphasized in~\cite{luo2024general}, the main challenge of this approach lies in the ambiguity of computing $\bm E(\cdot)$. The canonical pose space induced by PCA contains $48$ equivalent variants, corresponding to $3!$ possible eigenvalue orderings and $2^3$ eigenvector sign choices. This raises the key question:

\emph{How can we guarantee the uniqueness of $\bm E(\cdot)$ given any transformation of a point cloud $x$?}

We start similarly to~\cite{luo2024general}, considering $\bm \Lambda(\cdot)$ in \eqref{eq:PCA} in descending order, i.e., $\lambda_1>\lambda_2>\lambda_3\ge 0$ such that the cardinality of the canonical pose space is reduced from $48$ to $8$. After fixing the eigenvalue ordering, the remaining ambiguity lies in the signs of the eigenvectors. We resolve this challenge by proposing a novel \emph{asymmetry score} disambiguation approach. Specifically, we define a score function $\phi:\mathbb{R}^3 \times \mathbb{R}^{n\times 3}\to\mathbb{R}$, which, given a direction $\bm v$ and a centered point cloud $\bar{x}$, measures whether the mass of the cloud lies predominantly in the positive or negative half-space defined by $\bm v$.
Intuitively, $\phi$ assigns a sign to each candidate eigenvector by testing whether more points (or weighted points) project positively or negatively along that axis. As a result, this score function $\phi$ is required to satisfy
\begin{align}
\phi(\bm v,\{x(i)+\bm \iota ^\top\}) &= \phi(\bm v,\{x(i)\}), \ \text{(translation invariance)} \tag{C1} \label{eq:C1}\\
\phi(\bm R \bm v,\{x(i) \bm R^{\top}\}) &= \phi(\bm v,\{x(i)\}), \ \text{(rotation invariance)} \tag{C2} \label{eq:C2}\\
\phi(- \bm v,\{x(i)\}) &= -\phi(\bm v,\{x(i)\}), \ \text{(oddness)} \tag{C3} \label{eq:C3}
\end{align}
and examples include
\[
\phi(\bm v, x) = \sum_i \operatorname{sign}\!\langle \bar{x}(i), \bm v\rangle,
\ \text{or} \ \sum_i \|\bar{x}(i)\|^2 \operatorname{sign}\!\langle \bar{x}(i), \bm v\rangle.
\]

Thus, the signs of the first two eigenvectors are fixed consistently given the nonzero scores
\[
\sigma_1(\bm e_1, \bar{x}) = \operatorname{sign}\phi(\bm e_1, \bar{x}), 
\qquad
\sigma_2(\bm e_2, \bar{x}) = \operatorname{sign}\phi(\bm e_2, \bar{x}),
\]
while the third is determined by the right-hand rule
\begin{equation}
    \sigma_3^+(\bar{x}) := \det(\bm E(\bar{x})) \, \sigma_1(\bm e_1, \bar{x})\,\sigma_2(\bm e_2, \bar{x}),
    \label{eq:sigma_3}
\end{equation}
ensuring $\widehat{\bm E}(\bar{x}) \in SO(3)$ since
\begin{equation}
    \widehat{\bm E}(\bar{x}) := \bm E(\bar{x})\,\mathrm{diag}(\sigma_1, \sigma_2, \sigma_3^+).
    \label{eq:canonical_pose}
\end{equation}

Once computed $\widehat{\bm E}(\bar{x})$, the canonical pose is obtained by re-expressing the centered point cloud $\bar{x}$ in this frame:
\[
\PPC(x) = \bar{x} \widehat{\bm E}(\bar{x}) \in \mathbb{R}^{n \times 3}.
\]
In conclusion, uniqueness of the eigenbases $\bm{E}(\cdot)$ requires two main conditions: a simple spectrum $\lambda_1>\lambda_2>\lambda_3$, which fixes the eigenvector ordering, and a \emph{non-degenerate} score function $\phi$, i.e., $\phi(\bm e_1, \bar{x})\neq 0$ and $\phi(\bm e_2, \bar{x})\neq 0$ which satisfies conditions \eqref{eq:C1}, \eqref{eq:C2} and \eqref{eq:C3}. Under these conditions, generally true for real-world point clouds, $\PPC(x)$ is uniquely defined and ensure invariance to any rigid transformation of $x$. We theoretically justify this claim in the following paragraph.

\paragraph{Theoretical Results} Before introducing our invariance results, we recall a fundamental property of the covariance operator in \eqref{eq:PCA}: it is unaffected by translations, and transforms equivariantly under rotations. We formalize this statement in the following lemma

\begin{lemma}[Translation invariance and rotation equivariance]
\label{lemma:1}
Let $\mathcal{E}(x) = \{\bm E(x)\,\mathrm{diag}(\pm1,\pm1,\pm1)\}$ and 
$\mathcal{E}(x') = \{\bm E(x')\,\mathrm{diag}(\pm1,\pm1,\pm1)\}$ denote the sets of orthogonal eigenvector bases obtained from point sets $x$ and $x'$ respectively. Assume a simple spectrum $\lambda_1>\lambda_2>\lambda_3\ge 0$. If $x' = x \bm R^{\top} + \bm 1_n \bm \iota^\top$ with $\bm R\in SO(3)$, then the eigenbases transform by $\mathcal{E}(x')=\bm R\mathcal{E}(x)$.
\begin{proof}
    Let $\bar{x}$ and $\bar{x}'$ be the centered clouds of $x$ and $x'$.
    Since centering removes translations, $\bar{x}'=\bar{x}\,\bm R^\top$. Hence the covariances satisfy
    \(
    \Sigma(x')=\tfrac{1}{n-1}(\bar{x}')^\top \bar{x}'
    = \tfrac{1}{n-1}(\bm R \bar{x}^\top)(\bar{x}\bm R^\top)
    = \bm R\Sigma(x)\bm R^\top.
    \)
    If $\Sigma(x)\,\bm e_k=\lambda_k \bm e_k$, then
    \(
    \Sigma(x')(\bm R \bm e_k) 
    = \bm R\Sigma(x)\bm R^\top (\bm R \bm e_k)
    = \bm R\Sigma(x)\bm e_k
    = \lambda_k (\bm R \bm e_k),
    \)
    so $\bm R \bm e_k$ is an eigenvector of $\Sigma(x')$ with the same eigenvalue. Accounting for the (independent) sign flips, the set of PCA bases transforms as $\mathcal{E}(x')=\bm R\mathcal{E}(x)$.
\end{proof}
\end{lemma}

Lemma~\ref{lemma:1} established that PCA eigenvectors transform equivariantly under rigid motions, but this result still leaves the sign ambiguity unresolved. Lemma~\ref{lemma:2} shows that once eigenvector signs are fixed using the asymmetry score function $\phi$, the disambiguated eigenbasis inherits the same property: under any rigid motion, the disambiguated eigenbasis of the transformed point cloud is exactly the rotated version of the original disambiguated eigenbasis. This ensures that our canonical frame is well defined and rotation-consistent.

\begin{lemma}[Sign consistency]
\label{lemma:2}
Assume a simple spectrum $\lambda_1>\lambda_2>\lambda_3$ and that $\phi$ is non-degenerate on the top two axes: $\phi(\bm e_1,\bar{x})\neq 0$, $\phi(\bm e_2,\bar{x})\neq 0$. Define $\sigma_k(\bm e_k,\bar{x})=\operatorname{sign}\phi(\bm e_k,\bar{x})$ for $k\in\{1,2\}$, $\sigma_3^+(\bar{x})$ as in \eqref{eq:sigma_3}, and set $\widehat{\bm E}(x)$ as in \eqref{eq:canonical_pose}. If $x' = x\,\bm R^\top + \bm 1_n\,\bm \iota^\top$ with $\bm R\in SO(3)$, then $\widehat{\bm E}(x') \;=\; \bm R\,\widehat{\bm E}(x)$.
\begin{proof}
By Lemma~\ref{lemma:1}, if $\bm e_k$ is an eigenvector of $\Sigma(x)$ then $\bm R\bm e_k$ is an eigenvector of $\Sigma(x')$. 
Since $\bar{x}'=\bar{x}\bm R^\top$, the invariances of $\phi$ imply $\phi(\bm R\bm e_k,\bar{x}') = \phi(\bm e_k,\bar{x})$, and $\phi(-\bm R\bm e_k,\bar{x}') = -\phi(\bm R\bm e_k,\bar{x})$. Thus $\sigma_k(\bm R\bm e_k,\bar{x}')=\sigma_k(\bm e_k,\bar{x})$ for $k=1,2$, and
\[
\widehat{\bm e}_k' 
= \sigma_k(\bm R\bm e_k,\bar{x}')\,(\bm R\bm e_k) 
= \sigma_k(\bm e_k,\bar{x})\,(\bm R\bm e_k) 
= \bm R\,\widehat{\bm e}_k.
\]

For the third axis we apply the right-hand rule. Since $\bm R\in SO(3)$ we have $\det(\bm R)=1$, so $\det(\bm E') = \det(\bm R\bm E) = \det(\bm E)$. Therefore
\begin{align*}
\sigma_3^+(\bar{x}') = \det(\bm E')\,\sigma_1(\bm R\bm e_1,\bar{x}')\,\sigma_2(\bm R\bm e_2,\bar{x}') \\
= \det(\bm E)\,\sigma_1(\bm e_1,\bar{x})\,\sigma_2(\bm e_2,\bar{x}) = \sigma_3^+(\bar{x}).
\end{align*}
It follows that $\widehat{\bm e}_3' = \sigma_3^+(\bar{x}')\,(\bm R\bm e_3) = \sigma_3^+(\bar{x})\,(\bm R\bm e_3) = \bm R\,\widehat{\bm e}_3$. Stacking the columns gives $\widehat{\bm E}(x') = \bm R\,\widehat{\bm E}(x)$.
\end{proof}
\end{lemma}

Lemmas~\ref{lemma:1} and~\ref{lemma:2} establish that PCA eigenvectors transform equivariantly under rigid motions and that our sign-disambiguation procedure preserves this property. Together, these results imply the following Theorem~\ref{thm:rigid_invariance}, which shows that the canonical pose representation $\PPC(x)$ is invariant under any rigid transformation of the input point cloud. In other words, translating or rotating the cloud does not affect its canonicalized representation.

\begin{theorem}[Rigid-motion invariance on the non-degenerate simple-spectrum set]
\label{thm:rigid_invariance}
Assume $\lambda_1>\lambda_2>\lambda_3$ and that $x$ is non-degenerate with respect to $\phi$. For any rigid transformation $x' = x\,\bm R^\top + \bm 1_n \bm \iota^\top$ where $\bm R \in SO(3)$, the canonical pose is invariant: $\PPC(x') = \PPC(x)$.
\begin{proof}
By Lemma~\ref{lemma:2}, $\widehat{\bm E}(x') = \bm R \widehat{\bm E}(x)$. Since centering removes translations, we have $\bar{x}' = \bar{x}\bm R^\top$. Therefore, $\PPC(x') = \bar{x}'\widehat{\bm E}(\bar{x}') = (\bar{x}\bm R^\top)(\bm R \widehat{\bm E}(\bar{x})) = \bar{x}\,\widehat{\bm E}(\bar{x}) = \PPC(x)$.
\end{proof}
\end{theorem}

\begin{remark}[Degeneracies]
The uniqueness guarantee breaks down in two cases: $(i)$ when the spectrum is not simple (e.g.\ $\lambda_1=\lambda_2$), in which case the eigenbasis is not uniquely defined, and $(ii)$ when the asymmetry score vanishes for one of the first two axes, $\phi(\bm e_k,x)=0$ for $k\in{1,2}$, resulting in unresolved sign ambiguity. In such degenerate cases, the invariance property stated in Theorem~\ref{thm:rigid_invariance} no longer holds. However, these situations are rare in practice, as real-world centered point clouds typically exhibit sufficient asymmetry and noise to avoid exact degeneracies.
\end{remark}

\subsection{PPC in RL}
\label{subsec:PPC+PPRL}

We now turn to the RL setting and summarize how PPC can be integrated into PC-RL to guarantee invariance under rigid transformations. Figure~\ref{fig:pca_diambiguation} illustrates the full pipeline: voxelization and FPS first regularize the raw point cloud, reducing redundancy while preserving geometric coverage; PCA then defines a canonical basis; and the asymmetry score $\phi$ deterministically fixes axis directions, enforcing a consistent right-handed frame. By preprocessing observations with PPC, the downsampled input to the policy becomes invariant to translations and rotations, eliminating point cloud pose as a nuisance factor. This contrasts with augmentation-heavy methods, where the policy itself must learn invariances from data, often at the cost of higher sample complexity.

In our experiments, we associate PPC with PPRL, as it represents the state-of-the-art for PC-RL. Following the PPRL pipeline (Section~\ref{sec:preliminaries}), FPS is again employed to sample $m$ centroids from which $k$ local patches are extracted and tokenized. These tokens are then used to learn an embedding space $\mathcal{Z}$, which serves as the input to the downstream RL pipeline. In Section~\ref{sec:experiments} we show that PPC substantially improves PPRL zero-shot capabilities under viewpoint shifts. This modular design makes PPC compatible with any point-cloud-based agent.

\section{Experiments}
\label{sec:experiments}

We design our experiments to answer two core questions: $(i)$ can PPC enable robust zero-shot generalization in RL when agents face novel camera viewpoints at evaluation, and $(ii)$ does PPC-conferred translation and rotation invariance translate to approximate camera pose invariance for noisy real-world point clouds?

To address these, we evaluate PPC in two complementary settings. The first is a set of zero-shot RL experiments across diverse control tasks, where training occurs under a fixed viewpoint and evaluation introduces unseen camera poses. This stresses PPC’s ability to provide viewpoint invariance without hindering policy learning. The second focuses on real-world tabletop scenes captured by an RGB-D camera, providing a challenging test of stability under noise and perspective shifts. Together, these experiments assess both the practicality and guarantees of PPC across synthetic and real domains.

\subsection{Zero-shot in PC-RL}
\begin{figure}[!t]
    \centering
    \smallskip
    \smallskip
    \begin{subfigure}[t]{0.32\columnwidth}
        \centering
        \includegraphics[width=0.9\columnwidth]{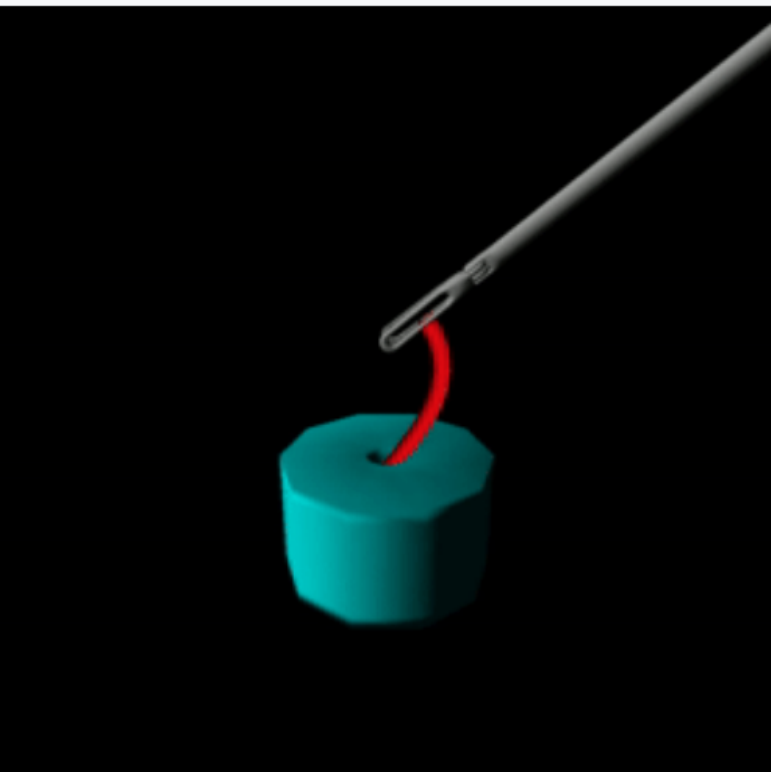}
        \caption{ThreadInHole}
        \label{fig:thread_in_hole}
    \end{subfigure}
    \begin{subfigure}[t]{0.32\columnwidth}
        \centering
        \includegraphics[width=0.9\columnwidth]{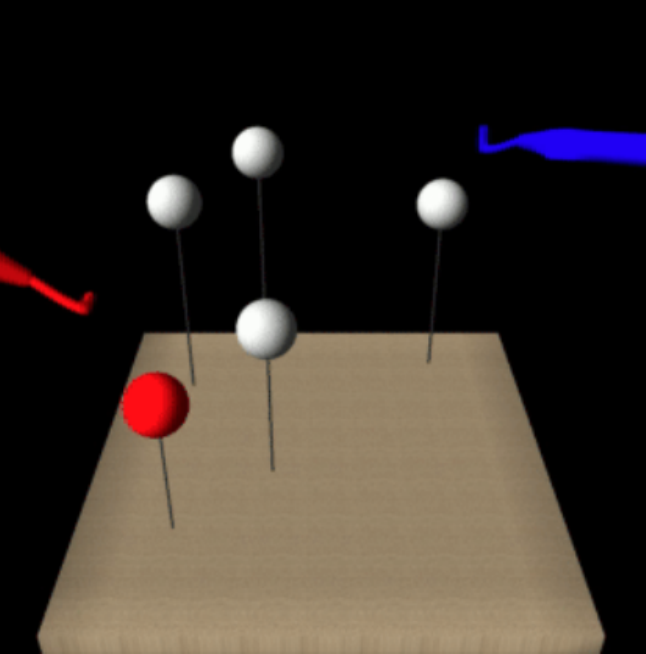}
        \caption{DeflectSpheres}
        \label{fig:deflect_spheres}
    \end{subfigure}
    \begin{subfigure}[t]{0.32\columnwidth}
        \centering
        \includegraphics[width=0.9\columnwidth]{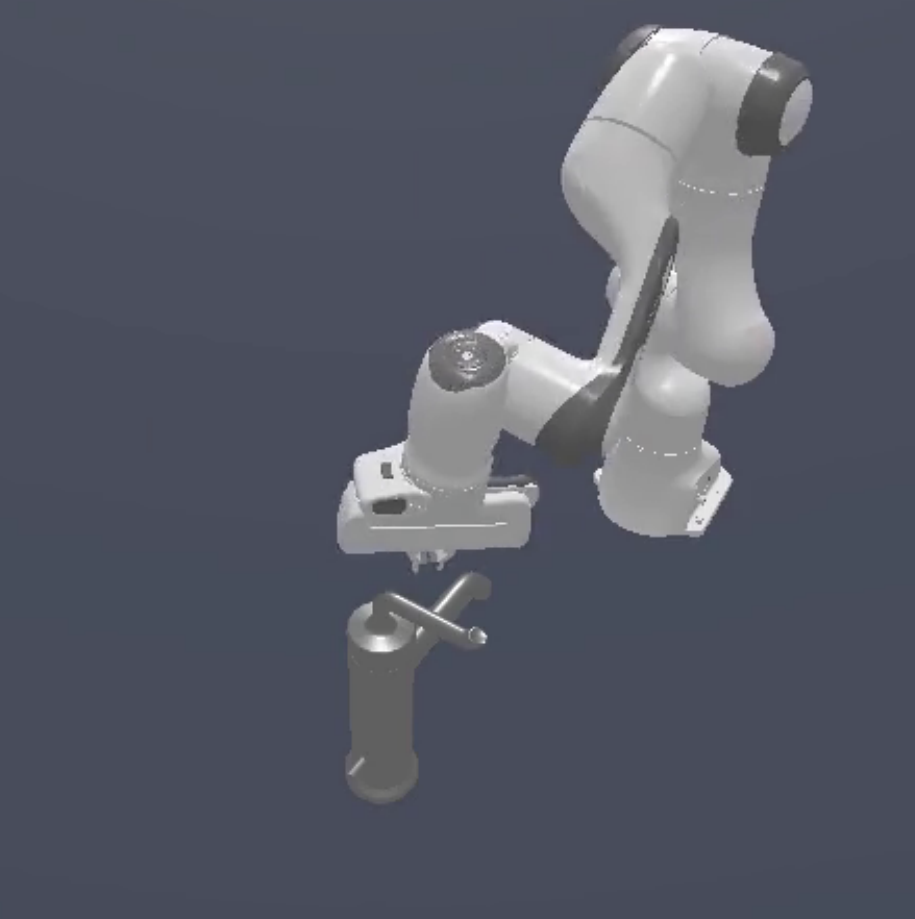}
        \caption{TurnFaucet}
        \label{fig:turn_faucet}
    \end{subfigure}
    \caption{Environments for zero-shot generalization in PC-RL.}
    \label{fig:fov_robustness}
\end{figure}

\begin{figure}[!h] 
    \centering 
    \smallskip
    \smallskip
    \begin{subfigure}[b]{1\columnwidth} 
        \centering
        \includegraphics[width=\columnwidth]{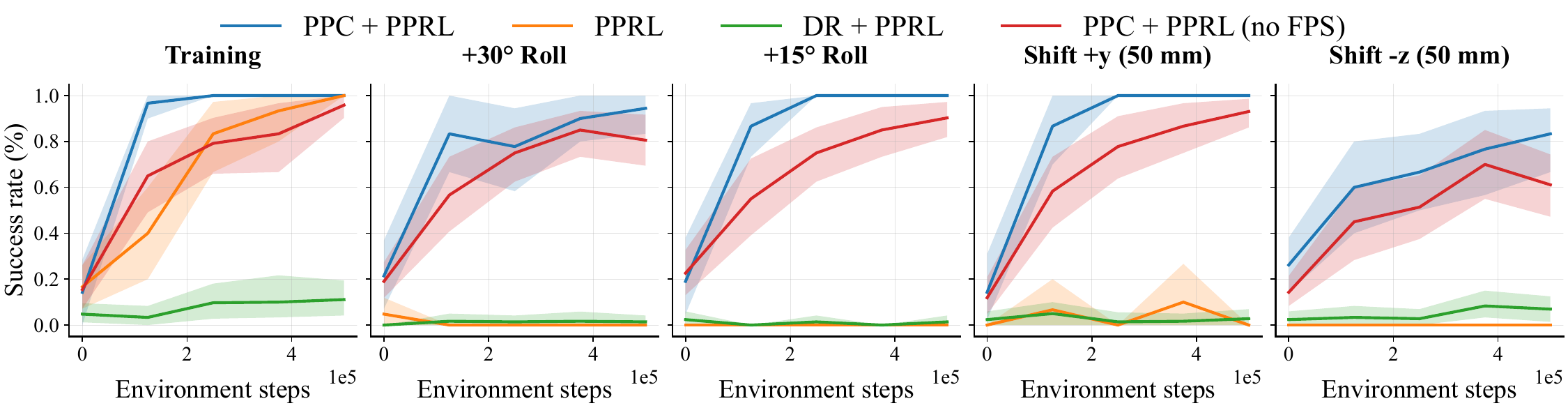} 
        \caption{Evaluation for ThreadInHole in Fig.~\ref{fig:thread_in_hole}}
        \vspace{0.1cm}
        \label{subfig:Thread_in_hole_evaluation}
    \end{subfigure}
    \begin{subfigure}[b]{1\columnwidth} 
        \centering
        \includegraphics[width=\columnwidth]{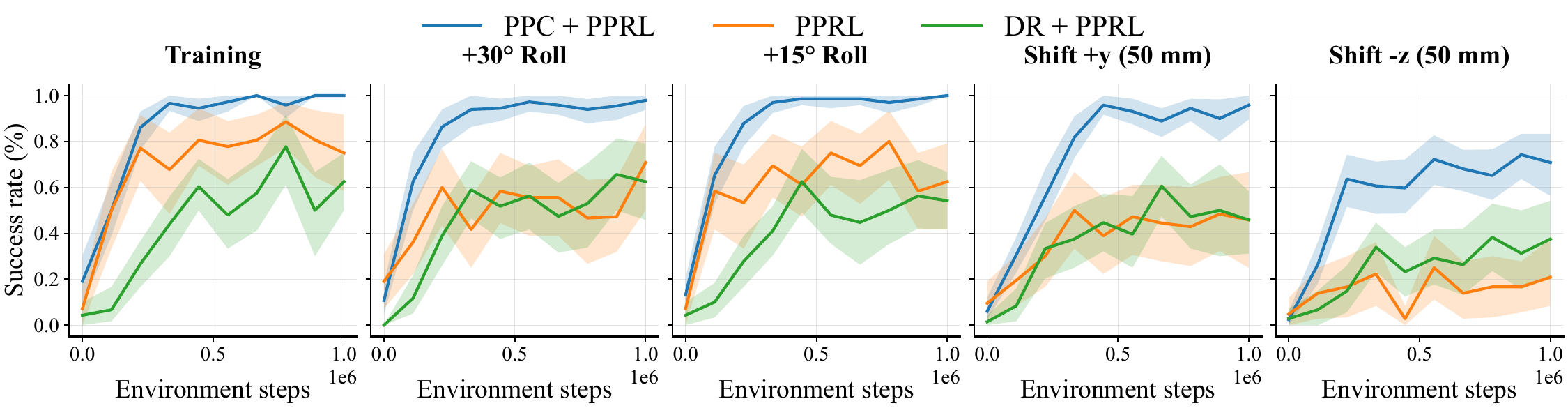} 
        \caption{Evaluation for DeflectSpheres in Fig.~\ref{fig:deflect_spheres}}
        \vspace{0.1cm}
        \label{subfig:Deflect_sphere_evaluation}
    \end{subfigure}
    \begin{subfigure}[b]{1\columnwidth} 
        \centering
        \includegraphics[width=\columnwidth]{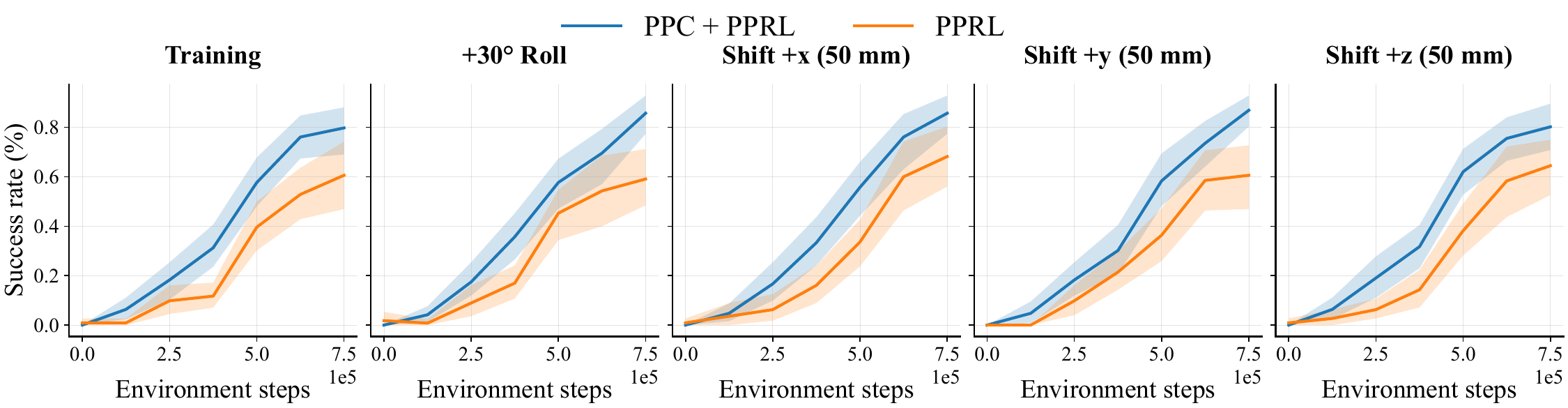}
        \caption{Evaluation for TurnFaucet in Fig.~\ref{fig:turn_faucet}}
        \vspace{0.1cm}
        \label{subfig:turn_faucet_evaluation}
    \end{subfigure}
    \caption{Summary of zero-shot generalization experiments in PC-RL. For each environment, the first plot depicts the algorithms' performance during training, while the others evaluate algorithms for zero-shot generalization under camera viewpoint mismatch. We run ThreadInHole for $5 \times10^5$ steps, DeflectSpheres for $10^6$ steps, and TurnFaucet for $7.5 \times 10^5$ steps, all for $6$ seeds. We evaluate over five episodes every $100$k steps, and plot means with 95\% bootstrapped confidence intervals across episodes and seeds. We use the auxiliary loss in PPRL for all tasks except DeflectSpheres, following the observation by Gyenes et al.~\cite{gyenes2024pointpatchrl} that it reduces performance in this setting.
    }
    \vspace{-0.4cm}
    \label{fig:rl_experiments}
\end{figure}


We evaluate \textit{PPC+PPRL} against standard \textit{PPRL}~\cite{gyenes2024pointpatchrl} and include two ablations: \textit{Domain Randomization with PPRL (DR+PPRL)} and \textit{PPC+PPRL without FPS}, and results are summarized in Fig.~\ref{fig:rl_experiments}. Experiments are conducted across two robot control suites: ManiSkill~\cite{gu2023maniskill2} and SofaEnv~\cite{JMLR:v24:23-0207}. From ManiSkill, we consider TurnFaucet (Fig.~\ref{fig:turn_faucet}), where a robot has to rotate a sink handle. From SofaEnv, designed for robot-assisted laparoscopic surgery, we evaluate ThreadInHole (Fig.~\ref{fig:thread_in_hole}), where a robotic arm has to thread a needle through a fixed hole, and DeflectSpheres (Fig.~\ref{fig:deflect_spheres}), where the agent deflects spheres using a correspondingly colored arm. While DeflectSpheres tests color reasoning, color cues are also present but non-essential in TurnFaucet and ThreadInHole.

Across tasks, \textit{PPC+PPRL} consistently achieves strong zero-shot performance, outperforming all baselines. In SofaEnv (Fig.~\ref{subfig:Thread_in_hole_evaluation} and \ref{subfig:Deflect_sphere_evaluation}), our algorithm yields notable robustness gains, with minimal degradation from training to evaluation. This robustness stems from PPC’s translation- and rotation-invariant properties, which align point cloud representations across viewpoints. 
For TurnFaucet (Fig.~\ref{subfig:turn_faucet_evaluation}), agents receive both vector state and point cloud inputs. Here, performance of \textit{PPC+PPRL} more closely matches standard \textit{PPRL}, suggesting the agent relies primarily on the vector state over the point cloud. Importantly, PPC does not interfere with learning in this setting; on the contrary, it improves overall training performance due to the stabilizing effect of its downsampling steps.

Our ablation study highlights complementary findings. Note that we perform this ablation only on SofaEnv environments due to computational constraints. For \textit{DR+PPRL}, domain randomization is applied by randomly shifting and rotating the camera, with evaluation at the $80$-th percentile of the randomization space. Initial trials showed that full random rotations after centering the point cloud inhibited convergence, motivating our choice of camera-pose variation. These experiments emphasize how, when convergence is achieved, \textit{DR+PPRL} improves robustness to camera viewpoints, as we observe minimal differences between training and evaluation. However, the added complexity due to randomization prevents convergence in ThreadInHole (Fig.~\ref{subfig:Thread_in_hole_evaluation}) and yields suboptimal performance in DeflectSpheres (Fig.~\ref{subfig:Deflect_sphere_evaluation}). By contrast, Fig.~\ref{subfig:Thread_in_hole_evaluation} shows that the absence of FPS decreases performance in both training and evaluation compared to \textit{PPC+PPRL}. This result emphasizes the importance of FPS in regularizing spatial point coverage. 

Taken together, our results demonstrate that \textit{PPC+PPRL} not only achieves robust alignment between training and evaluation but also improves learning across environments, underscoring the substantial advantages of PPC in PC-RL.

\subsection{PPC on Real-World Point Clouds}

\begin{figure}[!t]
\smallskip
\smallskip
    \begin{subfigure}[t]{\columnwidth}
        \centering
        \includegraphics[width=0.95\columnwidth]{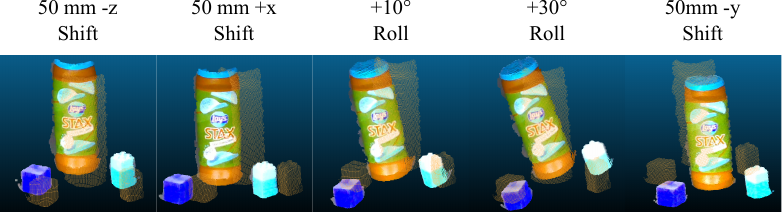}
        \caption{Scene 1}
        \vspace{0.2cm}
        \label{fig:sample_mismatches}
    \end{subfigure}
    \begin{subfigure}[t]{\columnwidth}
        \centering
        \includegraphics[width=0.95\columnwidth]{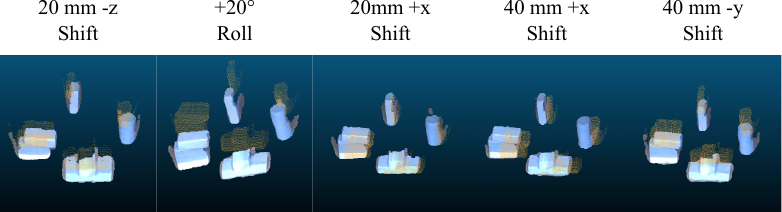}
        \caption{Scene 2}
    \label{fig:sample_mismatches}
    \end{subfigure}
    \caption{We visualize the selected test set of mismatches for the experiments in Table~\ref{tab:real_world_test}, showing the original view in orange and the perturbed view in RGB.}
    \vspace{-0.3cm}
    \label{fig:real_pc}
\end{figure}

To evaluate PPC beyond synthetic data, we applied it to real-world point clouds captured with an RGB-D camera. We recorded two tabletop scenes, each from a baseline camera pose and several alternative viewpoints, including shifts along each axis, rotations, and orbits (Fig.~\ref{fig:real_pc}). For each point cloud, we removed background points to retain only the objects of interest. This step can be readily automated in RL pipelines through segmentation.

Compared to experiments conducted in a simulated environment, real-world collection introduces additional challenges such as imperfect depth sensing and sensor noise. PPC must remain robust under these conditions without suffering from significant centroid drift or instability in the PCA axes. Perspective changes, uneven point distributions, and noisy depth values further increase the risk of disambiguation flips across views due to shifts in mass distribution of point clouds.

To investigate this effect, we evaluated three candidate score functions $\phi$ for the disambiguation step:
\begin{align*}
\phi_1(\bm v, x) &= \sum_i \| \bar{x}(i) \|^2 \mathrm{sign}\langle \bar{x}(i), \bm v \rangle\\
\phi_2(\bm v, x) &= \sum_i \| \bar{x}(i) \|^4 \mathrm{sign}\langle \bar{x}(i), \bm v \rangle\\
\phi_3(\bm v, x) &= \sum_i \| \bar{x}(i) \|^{-1} \mathrm{sign}\langle \bar{x}(i), \bm v \rangle.
\end{align*}

Furthermore, to evaluate the alignment between point clouds, we use the symmetric Chamfer distance:
\[
d_{\mathrm{CH}}(P,Q)=
\tfrac12\!\left(
\tfrac{1}{|P|}\sum_{\bm p\in P}\min_{\bm q\in Q}\|\bm p-\bm q\|_2+
\tfrac{1}{|Q|}\sum_{\bm q\in Q}\min_{\bm p\in P}\|\bm q-\bm p\|_2
\right),
\]
which computes the bidirectional average nearest-neighbor distance between clouds $P$ and $Q$, with lower values reflecting closer alignment of scene geometry between the target view and the baseline.

For each scene, we perform a 70-30 train-test split and apply 5-fold cross-validation on the training set to select the best $\phi$. During validation, all three candidate functions proved effective: each maintained a stable orientation across views in the disambiguation step and consistently reduced Chamfer distances relative to the original alignment.

Table~\ref{tab:real_world_test} reports test results common to $\phi_1$, $\phi_2$, and $\phi_3$, i.e., results where each $\phi_k$ performs in the same way, highlighting PPC’s ability to minimize viewpoint-shift differences $d_{CH}(\text{PPC}(\mathcal{U}(s)), \text{PPC}(\widetilde{\mathcal{U}}(s)))$. These findings suggest that in real-world settings, stability of a general $\phi$ is sufficient for deterministic disambiguation across viewpoints, and that different variants can be equally effective.

In summary, these experiments show that PPC transfers robustly from synthetic to real-world point clouds. Despite sensor noise and clutter, the method achieves stable alignment, and multiple $\phi$ choices yield equally strong performance without introducing sign instability. This robustness highlights the practicality of PPC in real sensing conditions, where reliability across diverse environments is essential.

\begin{table}[t]
\centering
\scriptsize
\begin{tabular}{c|c|c c c}
\toprule
Scene (Fig.~\ref{fig:real_pc}) & Mismatch & Raw & \textbf{PPC} \\
\midrule
\multirow{5}{*}{1} & back 50mm & 50.5& 2.4\\
 & right 50mm & 26.0 & 2.7 \\
 & rotate $10^\circ$ CW & 120.6 & 5.9\\
 & rotate $30^\circ$ CW & 117.2 & 6.5\\
 & up 50mm & 23.7 & 5.6\\
\midrule
\multicolumn{2}{c|}{Avg. (mm)} & 67.6 $\pm$ 48.0 &\textbf{4.6 $\pm$ 1.9}\\
\midrule
\multirow{5}{*}{2} & back 20mm & 33.4 &1.8\\
 & rotate $20^\circ$ CW & 33.2 & 4.4\\
 & right 20mm & 6.4 &5.7 \\
 & right 40mm & 12.1 &2.8\\
 &up 40mm & 36.1 &2.3\\
\midrule
\multicolumn{2}{c|}{Avg. (mm)} & 24.2 $\pm$ 13.9 &\textbf{3.4 $\pm$ 1.6}\\
\bottomrule
\end{tabular}

\caption{Summary for the experiments on the real-world point cloud in Fig.~\ref{fig:real_pc}. Values show the mean Chamfer distances on the test scene as compared to a default point cloud. PPC is observed to significantly improve alignment.}
\vspace{-0.4cm}
\label{tab:real_world_test}
\end{table}

\section{Conclusion}
\label{sec:conclusion}

In this work, we introduced PPC, a principled and lightweight approach for enforcing rotational and translational invariance in point cloud representations for robotic control. PPC resolves the challenge of axis sign ambiguity in PCA-based alignment through a novel disambiguation step that is theoretically grounded and stable across viewpoints.

Through our experiments, we demonstrate that PPC consistently enhances both training efficiency and zero-shot generalization in PC-RL. When evaluated under camera poses unseen during training, PPC enabled robust transfer, outperforming standard preprocessing pipelines and domain randomization. These gains highlight PPC’s practicality for seamless integration into existing RL frameworks. Beyond synthetic benchmarks, we validated PPC on real-world point clouds where occlusions, noise, and imperfect sensing often undermine PCA-based alignment. Across multiple scenes and perturbations, PPC produced stable canonical poses that significantly reduced Chamfer distances between baseline and shifted views. All three tested variants of the disambiguation function $\phi$ yielded reliable alignment, underscoring PPC’s robustness.

These results demonstrate that PPC is both theoretically principled and practically effective, bridging the gap between synthetic invariance guarantees and real-world robustness. We envision PPC as a building block for point cloud learning pipelines where cross-view consistency is essential, including robot learning, 3D scene understanding, and multi-view perception. Future work will extend PPC to real-world RL, explore settings with multiple or mobile cameras, incorporate segmentation to specify invariant scene objects, and investigate alternative PCA formulations.

\bibliographystyle{ieeetran}
\bibliography{mybib}

\begin{thebibliography}{10}
\providecommand{\url}[1]{#1}
\csname url@rmstyle\endcsname
\providecommand{\newblock}{\relax}
\providecommand{\bibinfo}[2]{#2}
\providecommand\BIBentrySTDinterwordspacing{\spaceskip=0pt\relax}
\providecommand\BIBentryALTinterwordstretchfactor{4}
\providecommand\BIBentryALTinterwordspacing{\spaceskip=\fontdimen2\font plus
\BIBentryALTinterwordstretchfactor\fontdimen3\font minus \fontdimen4\font\relax}
\providecommand\BIBforeignlanguage[2]{{%
\expandafter\ifx\csname l@#1\endcsname\relax
\typeout{** WARNING: IEEEtran.bst: No hyphenation pattern has been}%
\typeout{** loaded for the language `#1'. Using the pattern for}%
\typeout{** the default language instead.}%
\else
\language=\csname l@#1\endcsname
\fi
#2}}

\bibitem{mnih2015human}
V.~Mnih, K.~Kavukcuoglu, D.~Silver, A.~A. Rusu, J.~Veness, M.~G. Bellemare, A.~Graves, M.~Riedmiller, A.~K. Fidjeland, G.~Ostrovski, \emph{et~al.}, ``Human-level control through deep reinforcement learning,'' \emph{nature}, vol. 518, no. 7540, pp. 529--533, 2015.

\bibitem{pinto2018asymmetric}
L.~Pinto, M.~Andrychowicz, P.~Welinder, W.~Zaremba, and P.~Abbeel, ``Asymmetric actor critic for image-based robot learning,'' in \emph{14th Robotics: Science and Systems, RSS 2018}.\hskip 1em plus 0.5em minus 0.4em\relax MIT Press Journals, 2018.

\bibitem{chi2023diffusion}
C.~Chi, S.~Feng, Y.~Du, Z.~Xu, E.~Cousineau, B.~Burchfiel, and S.~Song, ``Diffusion policy: Visuomotor policy learning via action diffusion,'' in \emph{Robotics: Science and Systems}, 2023.

\bibitem{geles2024demonstrating}
I.~Geles, L.~Bauersfeld, A.~Romero, J.~Xing, and D.~Scaramuzza, ``Demonstrating agile flight from pixels without state estimation,'' \emph{Robotics Science and Systems online Proceedings}, no.~20, p. online, 2024.

\bibitem{yuan2023rl}
Z.~Yuan, S.~Yang, P.~Hua, C.~Chang, K.~Hu, and H.~Xu, ``Rl-vigen: A reinforcement learning benchmark for visual generalization,'' \emph{Advances in Neural Information Processing Systems}, vol.~36, pp. 6720--6747, 2023.

\bibitem{giammarino2024visually}
V.~Giammarino, J.~Queeney, and I.~C. Paschalidis, ``Visually robust adversarial imitation learning from videos with contrastive learning,'' \emph{arXiv preprint arXiv:2407.12792}, 2024.

\bibitem{ramazzina2025beyond}
A.~Ramazzina, V.~Giammarino, M.~El-Hariry, and M.~Bijelic, ``Beyond domain randomization: Event-inspired perception for visually robust adversarial imitation from videos,'' \emph{arXiv preprint arXiv:2505.18899}, 2025.

\bibitem{gyenes2024pointpatchrl}
B.~Gyenes, N.~Franke, P.~Becker, and G.~Neumann, ``Pointpatchrl--masked reconstruction improves reinforcement learning on point clouds,'' \emph{arXiv preprint arXiv:2410.18800}, 2024.

\bibitem{peri_point_2024}
\BIBentryALTinterwordspacing
S.~Peri, I.~Lee, C.~Kim, L.~Fuxin, T.~Hermans, and S.~Lee, ``Point {Cloud} {Models} {Improve} {Visual} {Robustness} in {Robotic} {Learners},'' 2024, version Number: 1. [Online]. Available: \url{https://arxiv.org/abs/2404.18926}
\BIBentrySTDinterwordspacing

\bibitem{ling2023efficacy}
Z.~Ling, Y.~Yao, X.~Li, and H.~Su, ``On the efficacy of 3d point cloud reinforcement learning,'' \emph{arXiv preprint arXiv:2306.06799}, 2023.

\bibitem{tobin2017domain}
J.~Tobin, R.~Fong, A.~Ray, J.~Schneider, W.~Zaremba, and P.~Abbeel, ``Domain randomization for transferring deep neural networks from simulation to the real world,'' in \emph{2017 IEEE/RSJ International Conference on Intelligent Robots and Systems (IROS)}.\hskip 1em plus 0.5em minus 0.4em\relax IEEE, 2017, pp. 23--30.

\bibitem{mehta2020active}
B.~Mehta, M.~Diaz, F.~Golemo, C.~J. Pal, and L.~Paull, ``Active domain randomization,'' in \emph{Conference on Robot Learning}.\hskip 1em plus 0.5em minus 0.4em\relax PMLR, 2020, pp. 1162--1176.

\bibitem{li2021closer}
F.~Li, K.~Fujiwara, F.~Okura, and Y.~Matsushita, ``A closer look at rotation-invariant deep point cloud analysis,'' in \emph{Proceedings of the IEEE/CVF International Conference on Computer Vision}, 2021, pp. 16\,218--16\,227.

\bibitem{luo2024general}
S.~Luo and W.~Gao, ``A general framework for rotation invariant point cloud analysis,'' in \emph{ICASSP 2024-2024 IEEE International Conference on Acoustics, Speech and Signal Processing (ICASSP)}.\hskip 1em plus 0.5em minus 0.4em\relax IEEE, 2024, pp. 3665--3669.

\bibitem{giammarinoadversarial}
V.~Giammarino, J.~Queeney, and I.~Paschalidis, ``Adversarial imitation learning from visual observations using latent information,'' \emph{Transactions on Machine Learning Research}.

\bibitem{hafner2019learning}
D.~Hafner, T.~Lillicrap, I.~Fischer, R.~Villegas, D.~Ha, H.~Lee, and J.~Davidson, ``Learning latent dynamics for planning from pixels,'' in \emph{Proceedings of the thirty-sixth International Conference on Machine Learning}.\hskip 1em plus 0.5em minus 0.4em\relax PMLR, 2019, pp. 2555--2565.

\bibitem{qi2017pointnet}
C.~R. Qi, H.~Su, K.~Mo, and L.~J. Guibas, ``Pointnet: Deep learning on point sets for 3d classification and segmentation,'' in \emph{Proceedings of the IEEE/CVF Conference on Computer Vision and Pattern Recognition}, 2017, pp. 652--660.

\bibitem{qi2017pointnet++}
C.~R. Qi, L.~Yi, H.~Su, and L.~J. Guibas, ``Pointnet++: Deep hierarchical feature learning on point sets in a metric space,'' \emph{Advances in Neural Information Processing Systems}, vol.~30, 2017.

\bibitem{Zhao_2021_ICCV}
H.~Zhao, L.~Jiang, J.~Jia, P.~H. Torr, and V.~Koltun, ``Point transformer,'' in \emph{Proceedings of the IEEE/CVF International Conference on Computer Vision (ICCV)}, October 2021, pp. 16\,259--16\,268.

\bibitem{doi:10.1142/S2811032324400010}
\BIBentryALTinterwordspacing
Y.~Pang, E.~H.~F. Tay, L.~Yuan, and Z.~Chen, ``Masked autoencoders for 3d point cloud self-supervised learning,'' \emph{World Scientific Annual Review of Artificial Intelligence}, vol.~02, p. 2440001, 2024. [Online]. Available: \url{https://doi.org/10.1142/S2811032324400010}
\BIBentrySTDinterwordspacing

\bibitem{Yu_2022_CVPR}
X.~Yu, L.~Tang, Y.~Rao, T.~Huang, J.~Zhou, and J.~Lu, ``Point-bert: Pre-training 3d point cloud transformers with masked point modeling,'' in \emph{Proceedings of the IEEE/CVF Conference on Computer Vision and Pattern Recognition (CVPR)}, June 2022, pp. 19\,313--19\,322.

\bibitem{chen2023pointgptautoregressivelygenerativepretraining}
\BIBentryALTinterwordspacing
G.~Chen, M.~Wang, Y.~Yang, K.~Yu, L.~Yuan, and Y.~Yue, ``Pointgpt: Auto-regressively generative pre-training from point clouds,'' 2023. [Online]. Available: \url{https://arxiv.org/abs/2305.11487}
\BIBentrySTDinterwordspacing

\bibitem{huang2021generalization}
W.~Huang, I.~Mordatch, P.~Abbeel, and D.~Pathak, ``Generalization in dexterous manipulation via geometry-aware multi-task learning,'' \emph{arXiv preprint arXiv:2111.03062}, 2021.

\bibitem{chen2022system}
T.~Chen, J.~Xu, and P.~Agrawal, ``A system for general in-hand object re-orientation,'' in \emph{Conference on Robot Learning}.\hskip 1em plus 0.5em minus 0.4em\relax PMLR, 2022, pp. 297--307.

\bibitem{wu2023learning}
Y.-H. Wu, J.~Wang, and X.~Wang, ``Learning generalizable dexterous manipulation from human grasp affordance,'' in \emph{Conference on Robot Learning}.\hskip 1em plus 0.5em minus 0.4em\relax PMLR, 2023, pp. 618--629.

\bibitem{liu2023frame}
M.~Liu, X.~Li, Z.~Ling, Y.~Li, and H.~Su, ``Frame mining: a free lunch for learning robotic manipulation from 3d point clouds,'' in \emph{Conference on Robot Learning}.\hskip 1em plus 0.5em minus 0.4em\relax PMLR, 2023, pp. 527--538.

\bibitem{qin2023dexpoint}
Y.~Qin, B.~Huang, Z.-H. Yin, H.~Su, and X.~Wang, ``Dexpoint: Generalizable point cloud reinforcement learning for sim-to-real dexterous manipulation,'' in \emph{Conference on Robot Learning}.\hskip 1em plus 0.5em minus 0.4em\relax PMLR, 2023, pp. 594--605.

\bibitem{fang2020rotpredictor}
J.~Fang, D.~Zhou, X.~Song, S.~Jin, R.~Yang, and L.~Zhang, ``Rotpredictor: Unsupervised canonical viewpoint learning for point cloud classification,'' in \emph{2020 International Conference on 3D Vision (3DV)}.\hskip 1em plus 0.5em minus 0.4em\relax IEEE, 2020, pp. 987--996.

\bibitem{zhang2019rotation}
Z.~Zhang, B.-S. Hua, D.~W. Rosen, and S.-K. Yeung, ``Rotation invariant convolutions for 3d point clouds deep learning,'' in \emph{2019 International Conference on 3D Vision (3DV)}.\hskip 1em plus 0.5em minus 0.4em\relax IEEE, 2019, pp. 204--213.

\bibitem{zhang2022riconv++}
Z.~Zhang, B.-S. Hua, and S.-K. Yeung, ``Riconv++: Effective rotation invariant convolutions for 3d point clouds deep learning,'' \emph{International Journal of Computer Vision}, vol. 130, no.~5, pp. 1228--1243, 2022.

\bibitem{xiao2021triangle}
C.~Xiao and J.~Wachs, ``Triangle-net: Towards robustness in point cloud learning,'' in \emph{Proceedings of the IEEE/CVF winter Conference on Applications of Computer Vision}, 2021, pp. 826--835.

\bibitem{zhao2022rotation}
C.~Zhao, J.~Yang, X.~Xiong, A.~Zhu, Z.~Cao, and X.~Li, ``Rotation invariant point cloud analysis: Where local geometry meets global topology,'' \emph{Pattern Recognition}, vol. 127, p. 108626, 2022.

\bibitem{xu2021sgmnet}
J.~Xu, X.~Tang, Y.~Zhu, J.~Sun, and S.~Pu, ``Sgmnet: Learning rotation-invariant point cloud representations via sorted gram matrix,'' in \emph{Proceedings of the IEEE/CVF International Conference on Computer Vision}, 2021, pp. 10\,468--10\,477.

\bibitem{kim2020rotation}
S.~Kim, J.~Park, and B.~Han, ``Rotation-invariant local-to-global representation learning for 3d point cloud,'' \emph{Advances in Neural Information Processing Systems}, vol.~33, pp. 8174--8185, 2020.

\bibitem{konda1999actor}
V.~Konda and J.~Tsitsiklis, ``Actor-critic algorithms,'' \emph{Advances in Neural Information Processing Systems}, vol.~12, 1999.

\bibitem{haarnoja2018soft}
T.~Haarnoja, A.~Zhou, K.~Hartikainen, G.~Tucker, S.~Ha, J.~Tan, V.~Kumar, H.~Zhu, A.~Gupta, P.~Abbeel, \emph{et~al.}, ``Soft actor-critic algorithms and applications,'' \emph{arXiv preprint arXiv:1812.05905}, 2018.

\bibitem{hassani2021escaping}
A.~Hassani, S.~Walton, N.~Shah, A.~Abuduweili, J.~Li, and H.~Shi, ``Escaping the big data paradigm with compact transformers,'' \emph{arXiv preprint arXiv:2104.05704}, 2021.

\bibitem{rusu20113d}
R.~B. Rusu and S.~Cousins, ``3d is here: Point cloud library (pcl),'' in \emph{2011 IEEE International Conference on Robotics and Automation (ICRA)}.\hskip 1em plus 0.5em minus 0.4em\relax IEEE, 2011, pp. 1--4.

\bibitem{gu2023maniskill2}
J.~Gu, F.~Xiang, X.~Li, Z.~Ling, X.~Liu, T.~Mu, Y.~Tang, S.~Tao, X.~Wei, Y.~Yao, \emph{et~al.}, ``Maniskill2: A unified benchmark for generalizable manipulation skills,'' \emph{arXiv preprint arXiv:2302.04659}, 2023.

\bibitem{JMLR:v24:23-0207}
\BIBentryALTinterwordspacing
P.~M. Scheikl, B.~Gyenes, R.~Younis, C.~Haas, G.~Neumann, M.~Wagner, and F.~Mathis-Ullrich, ``Lapgym - an open source framework for reinforcement learning in robot-assisted laparoscopic surgery,'' \emph{Journal of Machine Learning Research}, vol.~24, no. 368, pp. 1--42, 2023. [Online]. Available: \url{http://jmlr.org/papers/v24/23-0207.html}
\BIBentrySTDinterwordspacing

\end{thebibliography}

\end{document}